\documentclass[runningheads]{llncs}

\usepackage{eccv}

\usepackage{eccvabbrv}

\usepackage{graphicx}
\usepackage{booktabs}
\usepackage{parskip}

\usepackage[accsupp]{axessibility}  % Improves PDF readability for those with disabilities.

\usepackage[pagebackref,breaklinks,colorlinks,citecolor=eccvblue]{hyperref}
\usepackage{orcidlink}
\usepackage[dvipsnames]{xcolor}

\newcommand{\bx}{\textbf{\emph{x}}}

\newcommand{\bphi}{\boldsymbol{\phi}}
\newcommand{\beps}{\boldsymbol{\epsilon}}
\newcommand{\ba}{\textbf{\emph{a}}}
\newcommand{\bb}{\textbf{\emph{b}}}
\newcommand{\bc}{\textbf{\emph{c}}}

\newcommand{\bX}{\textbf{\emph{X}}}

\newcommand{\bI}{\textbf{\emph{I}}}
\newcommand{\bJ}{\textbf{\emph{J}}}

\newcommand{\bK}{\textbf{\emph{K}}}

\newcommand{\bR}{\textbf{\emph{R}}}

\newcommand{\bk}{\textbf{\emph{k}}}

\newcommand{\bS}{\textbf{\emph{S}}}

\newcommand{\bSigma}{\boldsymbol{\Sigma}}

\usepackage{float}
\usepackage{dblfloatfix}
\usepackage{nicematrix}

\begin{document}

\title{LOSTU: Fast, Scalable, and Uncertainty-Aware Triangulation} 

\titlerunning{LOSTU}

\author{Sebastien Henry\inst{1}\orcidlink{0000-0002-6303-5863} \and
John A. Christian\inst{1}\orcidlink{0000-0002-9522-5784}}

\authorrunning{Henry S. \& Christian J. A.}

\institute{$^1$Georgia Institute of Technology
\email{\{seb.henry, john.a.christian\}@gatech.edu}\\
\url{https://seal.ae.gatech.edu/} } %\and

\maketitle

\begin{abstract}
This work proposes a non-iterative, scalable, and statistically optimal way to triangulate called \texttt{LOSTU}. Unlike triangulation algorithms that minimize the reprojection ($L_2$) error, \texttt{LOSTU} will still provide the maximum likelihood estimate when there are errors in camera pose or parameters. This generic framework is used to contextualize other triangulation methods like the direct linear transform (DLT) or the midpoint. Synthetic experiments show that \texttt{LOSTU} can be substantially faster than using uncertainty-aware Levenberg-Marquardt (or similar) optimization schemes, while providing results of comparable precision. Finally, \texttt{LOSTU} is implemented in sequential reconstruction in conjunction with uncertainty-aware pose estimation, where it yields better reconstruction metrics.

%Triangulation algorithms often aim to minimize the reprojection ($L_2$) error, but this only provides the maximum likelihood estimate when there are no errors in the camera parameters or camera poses.  Although recent advancements have yielded techniques to estimate camera parameters accounting for 3D point uncertainties, most structure from motion (SfM) pipelines still use older triangulation algorithms. This work leverages recent discoveries to provide a fast, scalable, and statistically optimal way to triangulate called \texttt{LOSTU}. Results show that \texttt{LOSTU} consistently produces lower 3D reconstruction errors than conventional $L_2$ triangulation methods---often allowing \texttt{LOSTU} to successfully triangulate more points. Moreover, in addition to providing a better 3D reconstruction, \texttt{LOSTU} can be substantially faster than Levenberg-Marquardt (or similar) optimization schemes.
\end{abstract}

\captionsetup{type=figure}
\begin{center}
    \subfloat[Vesta from segment RC3b. \label{fig:true_picture}]{%
        \includegraphics[width=0.24\textwidth]{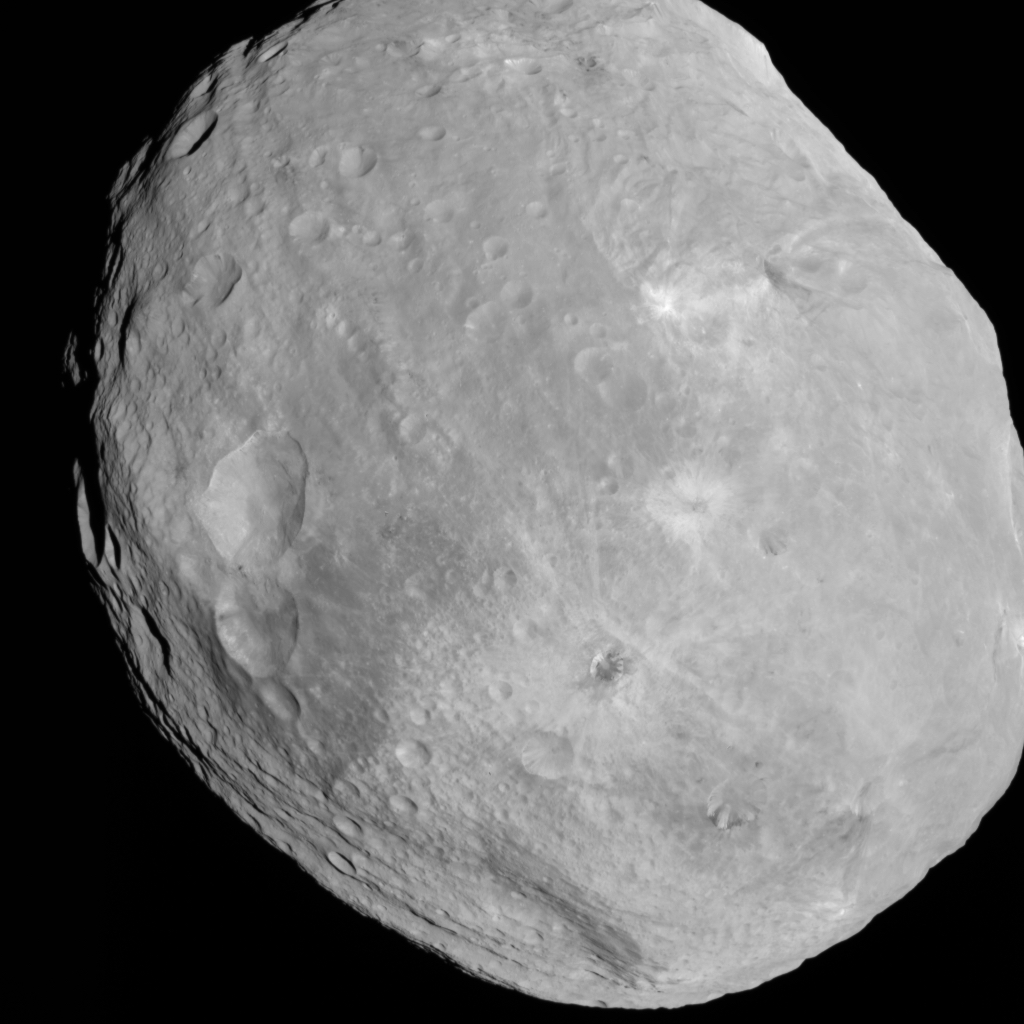}%
    } \hfill
    \subfloat[\texttt{EPnPU+DLT}.  \label{fig:vesta_DLT}]{%
        \includegraphics[width=0.24\textwidth]{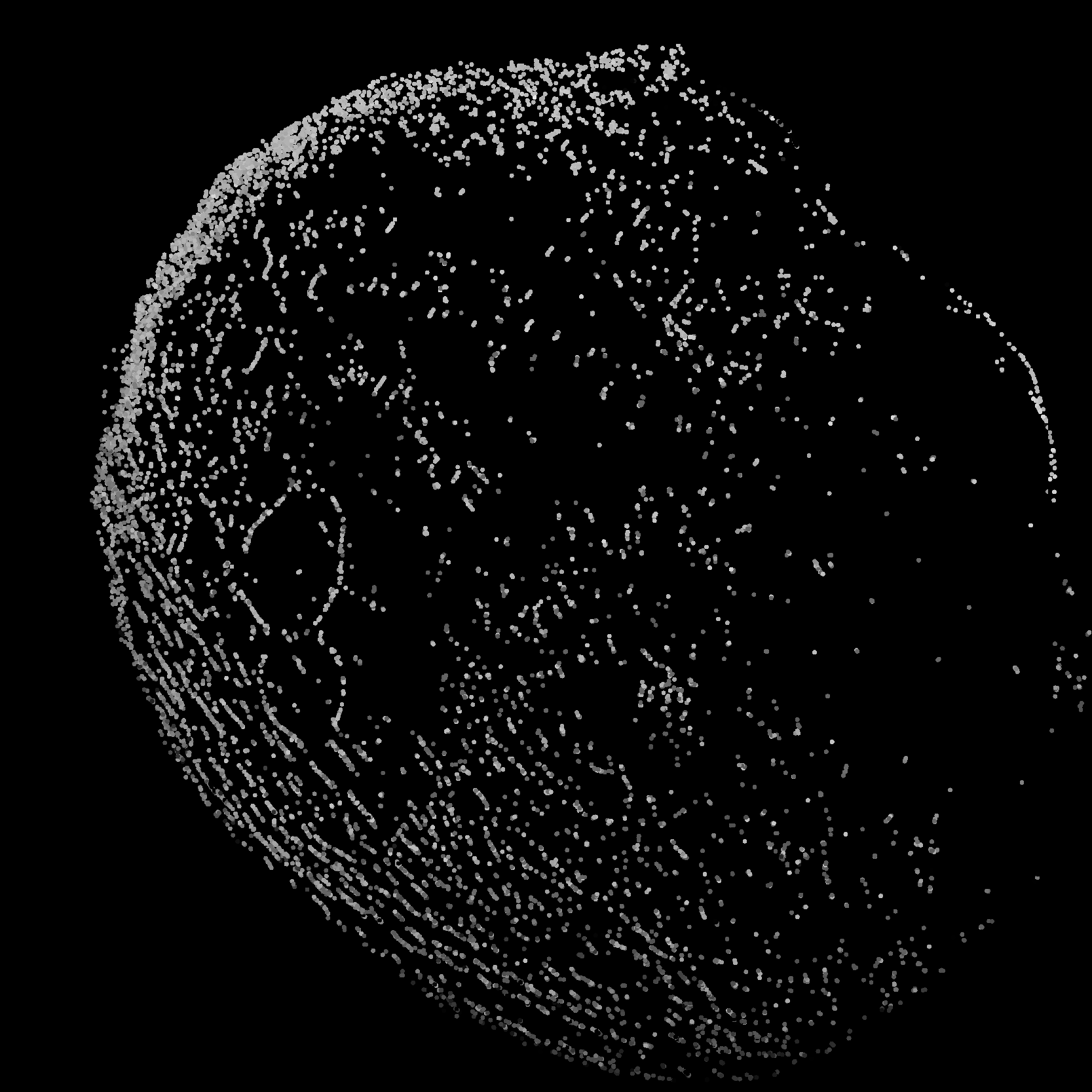}%
    } \hfill
    \subfloat[\texttt{EPnPU+LOSTU}. \label{fig:vesta_LOSTU}]{%
        \includegraphics[width=0.24\textwidth]{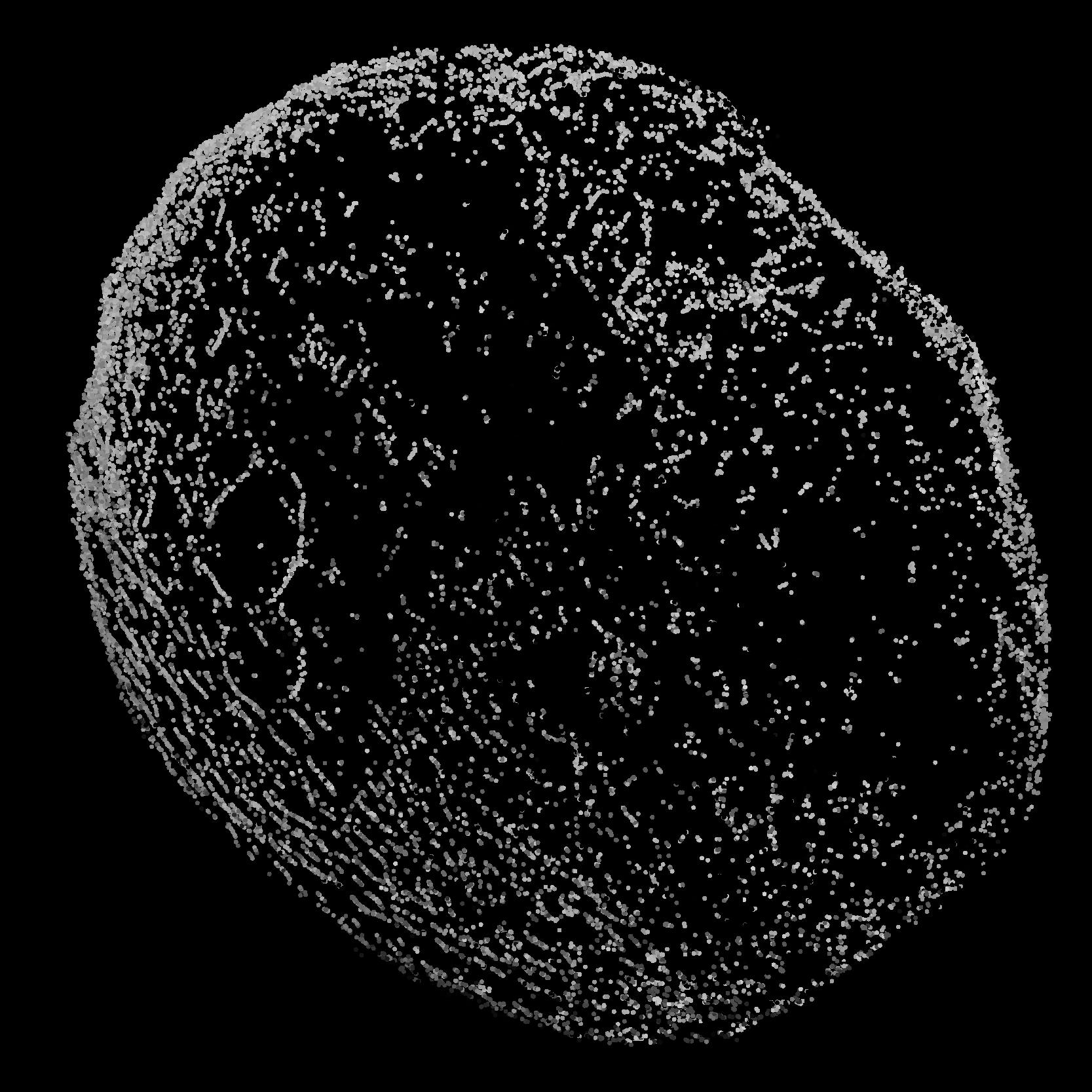}%
    } \hfill
    \subfloat[Ground truth+\texttt{LOST}. \label{fig:vesta_LOST}]{%
        \includegraphics[width=0.24\textwidth]{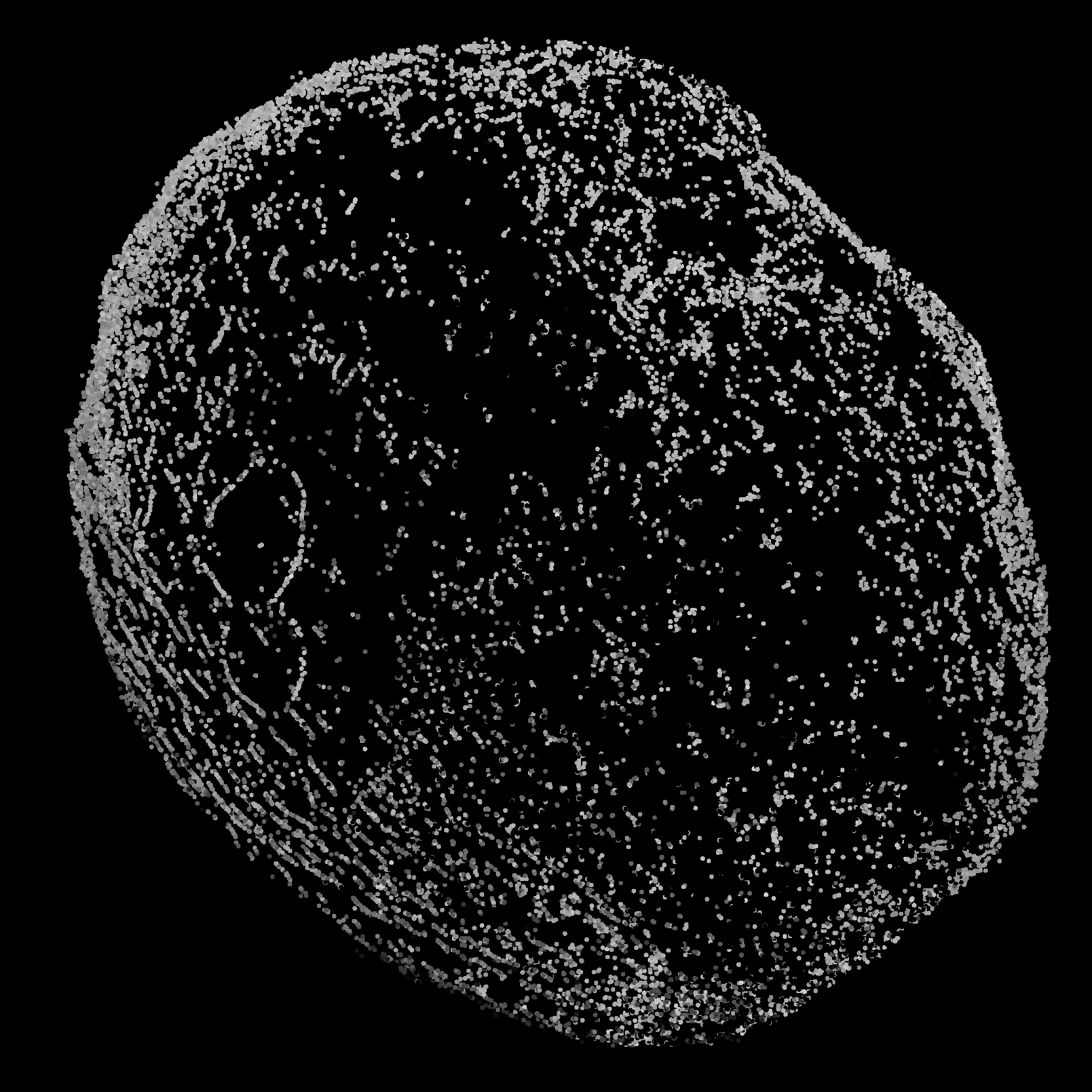}%   
    } \hfill \\
    \caption{Reconstruction of the asteroid Vesta from images of the RC3b phase of the Dawn mission. The images and ground truth are from the Astrovision dataset \cite{Driver:2023_AstroVision}. For the same constraint on the maximum covariance of a reconstructed point, the maximum likelihood triangulation method is able to reconstruct more points.}
\label{fig:vesta_reconstruction}
\end{center}

\section{Introduction}
\label{sec:intro}
Triangulation describes the task of localizing a point by the intersection of two---or more---lines-of-position (LOPs). In computer vision applications, these LOPs are usually referred as projection rays and originate from lines-of-sight (LOS) directions that are transformed from a camera frame into the world frame. Such situations arise naturally in image-based 3D reconstruction and navigation problems. Although practical triangulation algorithms have been around for hundreds of years \cite{Henry:2022}, it remains a problem of contemporary study.

Several important points are to be considered when choosing a triangulation method: its optimality, its scalability to multiple views, and its projective invariance property. It can be shown that under the assumption of Gaussian 2D noise, the solution minimizing the reprojection error provides the maximum likelihood estimate and is projective invariant \cite{hartley_zisserman_2004}. This type of approach is often named $L_2$ or \emph{optimal} triangulation, but this cost function is difficult to work with because it has multiple local minima \cite{Hartley:1997}---unlike $L_\infty$ triangulation \cite{Hartley:2004_Linf, Kahl:2008} that has a single local (and thus global) minimum. 

In 2-view triangulation, the optimal solution can be found by solving for the root of a sixth-order polynomial \cite{Hartley:1997}, and faster alternative methods exist \cite{Kanatani:2008, Lindstrom:2010}. Yet, the problem quickly becomes intractable as more views are added. It is, for example, necessary to find the roots of a 47th degree polynomial for 3 views \cite{Stewenius:2005}. Some notable work has also been done to speed up 3-view triangulation \cite{Byrod:2007, Kukelova:2013} but is not scalable to more points. Triangulation algorithms like the direct linear transform (DLT) \cite{hartley_zisserman_2004} and the midpoint method \cite{Hartley:1997} are popular due to their scalability to multiple views. These, or $L_\infty$ triangulation \cite{Lu:2007}, are often used to obtain an initial estimate that is then iteratively refined to minimize the reprojection error. However, the refinement can be slow and may not converge to the global optimum. Subsequent methods---like the linear optimal sine triangulation (LOST)---yield a closed-form, fast, and scalable solution to $L_2$ triangulation by weighting the DLT \cite{Henry:2022}. This framework has been shown to give statistically similar results to the common iterative optimal approach \cite{gtsam2023_losttriangulation}. 

Understanding the limitations of $L_2$ triangulation is essential. First, there are certain geometries (typically symmetric geometries) under which all triangulation methods perform similarly \cite{Hartley:1997, Henry:2022} and hence where optimal triangulation does not offer substantial performance gains. Furthermore, geometries involving low-parallax are a typical example where $L_2$ triangulation performs poorly \cite{Hartley:1997, lee:2019triangulation}. Regardless, $L_2$ triangulation can provide substantial improvements over sub-optimal methods in general scenarios that are not symmetric and don't have low parallax \cite{Hartley:1997, Henry:2022}.

Second, minimizing the reprojection error is only optimal under the assumptions of Gaussian 2D noise on the image measurements and perfect knowledge of the camera parameters. Reference~\cite{Nasiri:2023} argues that the midpoint method should be used over $L_2$ optimal triangulation in the SfM process because of uncertainties in the extrinsic camera parameters. However, the midpoint remains a method that inherently minimizes the wrong cost function. A more appropriate approach under camera parameter uncertainties is to modify the $L_2$ cost function in the iterative refinement process, as it is done in Ref.~\cite{Bedekar:1995_BayesianTriangulation, Haner:2012_nextBestView, lee:2020_robust}. Spurred by the challenges in spacecraft localization, the LOST algorithm has been used to analyze the relative importance of planetary uncertainty and spacecraft camera attitude uncertainties \cite{Henry:2023_LOSTU}, analytically. While the equations once again feature a linear system, no numerical simulation to validate that method has been done in Ref.~\cite{Henry:2023_LOSTU} since it was concluded that these effects were mostly negligible in spaceflight.

This paper carefully studies triangulation under camera pose uncertainty and address performance effects with traditional triangulation methods. By the geometric equivalence between intersection and resection \cite{Henry:2022}, we show that ``planetary position uncertainties'' become ``camera center uncertainties'' in a reconstruction scenario. We extend the framework in Refs.~\cite{Henry:2022, Henry:2023_LOSTU}, to introduce \texttt{LOSTU}: a general uncertainty-aware and non-iterative framework for triangulation. Existing studies, like Ref.~\cite{Nasiri:2023} and \cite{Haner:2012_nextBestView}, either lack uncertainty-aware optimal triangulation methods or use iterative refinement within a larger framework. Therefore, we perform simulations to experimentally validate \texttt{LOSTU} and the advantages of considering camera parameter uncertainties. These benefits are further demonstrated in a maximum likelihood sequential reconstruction pipeline.

\section{Linear optimal sine framework}
This work utilizes the pinhole camera model. For simplicity of notation, we denote the vector $\bk = [0, 0, 1]^T$. Then, the homogeneous pixel coordinate measurement of point $i$ in view $j$ is often represented as
\begin{equation}
    \bx_{ij} = \bK_j \frac{\bR^W_{C_j} \left( \bX_i  - \bc_j\right)}{\bk^T \bR^W_{C_j} \left( \bX_i  - \bc_j\right)},
\end{equation}
where $\bK_j$ is the camera calibration matrix, $\bx_{ij}$ is the 2D measurement of the object in homogeneous coordinates, $\bR^{W}_{C_j}$ is the rotation matrix from world frame to camera frame, $\bX_i$ is the 3D world position of the measured object, and $\bc_j$ is the 3D world position of the camera. The reader will recognize that, under perfect measurement, the measurement vector should be colinear to the line-of-sight
\begin{equation}
    \bK_j^{-1} \bx_{ij} \propto \rho_{ij} \frac{\bK_j^{-1} \bx_{ij}}{\| \bK_j^{-1} \bx_{ij} \|} = 
    \rho_{ij} \ba_{ij} = \bR^W_{C_j} \left( \bX_i  - \bc_j\right)  ,
\end{equation}
where $ \rho_{ij} =\| \bX_i - \bc_j \|$ is the range and $\ba_{ij}$ is the unit vector in the direction of the measurement in the camera frame. Thus, the cross product between those two vectors results in the zero vector (a vector writing of the law of sines). In practice, the 2D measurements are perturbed by noise, often assumed Gaussian, and this leads to the equation for the law of sines residual
\begin{align}
    \label{eq:residual}
        \beps_{ij} = \left[ \bK_j^{-1} \bx_{ij} \times \right] \bR^W_{C_j} (\bX_i - \bc_j ).
\end{align}
where $[\, \cdot \, \times]$ is the skew symmetric cross product matrix $\ba \times \bb = [\ba \times ] \bb$. The partials of $\beps_{ij}$ with respect to the 2D measurement are
\begin{equation}
    \label{eq:partial_2D}
    \bJ_{\bx_{ij}}  = \partial \beps_{ij} / \partial \bx_{ij} = - \left[ \bR^W_{C_i} (\bX_i - \bc_j ) \times \right] \bK_j^{-1}.
\end{equation}
Similarly, the partials with respect to the extrinsic camera parameters can be obtained as Ref.~\cite{Henry:2023_LOSTU}:
\begin{subequations}
\begin{align}
    \label{eq:partial_translation}
    &\bJ_{\bc_j} = \partial \beps_{ij} / \partial \bc_j = - \left[ \bK_j^{-1} \bx_{ij} \times \right] \bR^W_{C_j},\\
    \label{eq:partial_rotation}
    &\bJ_{\bphi_j} = \partial \beps_{ij} / \partial \bphi_j = \left[ \bK_j^{-1} \bx_{ij} \times \right]  \left[ \bR^W_{C_j} (\bX_i - \bc_j ) \times \right],
\end{align}
\end{subequations}
where $\bphi_j$ is the angle-vector description of a rotation perturbation in $\bR^W_{C_j}$. The partials with respect to the 3D world positions are
\begin{align}
\label{eq:partial_3Dpoint}
    &\bJ_{\bX_{i}}  = \partial \beps_{ij} / \partial \bX_{i} = \left[ \bK_j^{-1} \bx_{ij} \times \right] \bR^W_{C_j}.
\end{align}
Because the inverse of the calibration matrix has a simple closed-form \cite{Christian:2021_OPNAV_tutorial}, we can take the partials with respect to the desired intrinsic parameters as
\begin{align}
    \label{eq:partial_intrinsics}
    &\bJ_{\bK_j[l,m]} = \frac{\partial \beps_{ij}}{\partial \bK_j[l,m]} =  - \left[ \bR^W_{C_i} (\bX_i - \bc_j ) \times \right]  \frac{\partial \bK_j^{-1} }{\partial \bK_j[l,m]} \bx_{ij},
\end{align}
where $\bK_j[l,m]$ is the $l,m$-th entry of the calibration matrix.
The partials to other calibration parameters such as the radial distortion could also be taken into account by taking the appropriate Jacobians.

Assuming uncorrelated Gaussian noise models where $\bSigma_{(.)}$ represent the covariance matrix of $(.)$, \cref{eq:partial_2D,eq:partial_translation,eq:partial_rotation,eq:partial_3Dpoint,eq:partial_intrinsics} make it possible to project the individual parameter uncertainties onto the residual uncertainties as
\begin{equation} \label{eq:residual_cov}
    \bSigma_{\beps_{ij}} = \bJ_{\bphi_{j}} \bSigma_{\bphi_{j}} \bJ_{\bphi_{j}}^T + \bJ_{\bx_{ij}} \bSigma_{\bx_{ij}} \bJ_{\bx_{ij}}^T + \ldots .
\end{equation}
Denote the set of points visible in view $j$ as $\mathcal{V}_j$. The MLE is the solution that minimizes the cost function
\begin{equation} \label{eq:cost_function_prelim}
    J(\bK, \bR, \bc, \bx, \bX) = \sum_{j}\sum_{i\in\mathcal{V}_j} \beps_{ij}^T \bSigma_{\beps_{ij}}^{-1} \beps_{ij}.
\end{equation}
Due to the fact that the first cross product in \cref{eq:partial_2D} to \cref{eq:partial_intrinsics} have the same null space, the matrix $\bSigma_{\beps_{ij}}$ is not full rank and thus not invertible. The trick resides in observing that the null space of $\bSigma_{\beps_{ij}}$ naturally aligns with the residual of \cref{eq:residual}, and one can thus use the pseudo-inverse $\bSigma_{\beps_{ij}}^{\dagger}$ instead of the matrix inversion to rewrite \cref{eq:cost_function_prelim} as
\begin{equation} \label{eq:cost_function}
    J(\bK, \bR, \bc, \bx, \bX) = \sum_{j}\sum_{i\in\mathcal{V}_j} \beps_{ij}^T \bSigma_{\beps_{ij}}^{\dagger} \beps_{ij}.
\end{equation}

\subsection{Linear optimal sine triangulation with uncertainties (LOSTU)} \label{sec:LOST_intersection}
Denote the track as the set $\mathcal{T}_i$ that consists of all the views that see the $i$th point, $\mathcal{T}_i = \{j: i\in \mathcal{V}_j\}$. In the case of intersection, we seek to estimate the point $\bX_i$, and the cost function in \cref{eq:cost_function} becomes
\begin{equation} \label{eq:cost_function_3D}
    J(\bX_i) = \sum_{j\in\mathcal{T}_i} \beps_{ij}^T \bSigma_{\beps_{ij}}^{\dagger} \beps_{ij}
\end{equation}
where $\bSigma_{\beps_{ij}}$ is computed using \cref{eq:residual_cov}, but does not take the partial derivative at \cref{eq:partial_3Dpoint} into consideration. If an initial estimate for the 3D point exists, then it is straightforward to compute the Jacobians. When no \textit{a-priori} information is available, the covariance of the residual can still be computed. We can start by the acknowledging that
\begin{equation}
    \left[ \bR^W_{C_i} (\bX_i - \bc_j ) \times \right] = \rho_{ij} \left[ \ba_{ij} \times \right].
\end{equation}
Given that the camera centers are known (in intersection), the range $\rho_{ij} = \| \bX_i - \bc_j \|$ can be computed with the help of another measurement $\bx_{ij'}$ using the law of sines \cite{Henry:2022}
\begin{equation} \label{eq:approx_range}
    \rho_{ij} = \frac{\| \bR^W_{C_i} (\bc_j - \bc_{j'}) \times \ba_{ij}\|}{\|\ba_{ij} \times \ba_{ij'} \|}.
\end{equation}
The optimal point is the point satisfying $\partial J(\bX_i)/\partial \bX_i = \mathbf{0}$, which yields the final system to triangulate the position of the $i$th 3D point \cite{Henry:2022}
\begin{equation} \label{eq:LOST_Normal}
    \begin{aligned}
    \left(\sum_{j \in \mathcal{T}_i} \bR^{C_j}_W \left[ \bK_j^{-1} \bx_{ij} \times \right] \bSigma_{\beps_{ij}}^{\dagger} \left[ \bK_j^{-1} \bx_{ij} \times \right] \bR^W_{C_j} \right) \bX_i \\
    = \sum_{j \in \mathcal{T}_i} \bR^{C_j}_W \left[ \bK_j^{-1} \bx_{ij} \times \right] \bSigma_{\beps_{ij}}^{\dagger} \left[ \bK_j^{-1} \bx_{ij} \times \right] \bR^W_{C_j} \bc_j .
\end{aligned}
\end{equation}

Assuming isotropic 2D noise only ($\bSigma_{x_{ij}} = \sigma_{\bx_{ij}}^2\bI_{2\times2}$), the system can be further simplified by the QR factorization of the $\bSigma_{\beps_{ij}}^{\dagger}$. Denote the weights \cite{Henry:2022}
\begin{equation} \label{eq:LOST_weight}
    q_j = \frac{\| \bK_j^{-1} \bx_{ij} \|}{\bK_j^{-1}[0,0]\sigma_{\bx_{ij}} \rho_{ij}},
\end{equation}
where $\rho_{ij}$ can be computed with  \cref{eq:approx_range} and $K[0,0]$ is the first diagonal element of $K$. The system to solve can be rewritten as \cite{Henry:2022}
\begin{equation}
\begin{aligned} \label{eq:LOST_Ax=b}
    &\begin{bmatrix}
        q_1 \bS \left[ \bK_{j_1}^{-1} \bx_{i{j_1}} \times \right] \bR^W_{C_{j_1}} \\
        q_2 \bS \left[ \bK_{j_2}^{-1} \bx_{i{j_2}} \times \right] \bR^W_{C_{j_2}} \\
        \vdots \\
        q_n \bS \left[ \bK_{j_n}^{-1} \bx_{i{j_n}} \times \right] \bR^W_{C_{j_n}} \\
    \end{bmatrix} \bX_i =     \begin{bmatrix}
        q_1 \bS \left[ \bK_{j_1}^{-1} \bx_{i{j_1}} \times \right] \bR^W_{C_{j_1}} \bc_{j_1}\\
        q_2 \bS \left[ \bK_{j_2}^{-1} \bx_{i{j_2}} \times \right] \bR^W_{C_{j_2}} \bc_{j_2}\\
        \vdots \\
        q_n \bS \left[ \bK_{j_n}^{-1} \bx_{i{j_n}} \times \right] \bR^W_{C_{j_n}} \bc_{j_n} \\
    \end{bmatrix} 
\end{aligned}
\end{equation}
, where $\{j_1, \ldots j_n\} \in \mathcal{T}_i$ and $\bS = [\bI_{2\times2}, \mathbf{0}_{2\times1}]$. 

The expression in \cref{eq:LOST_weight} can help us understand when optimal triangulation matters most over the DLT. The DLT solution is obtained by solving the system \cref{eq:LOST_Normal} replacing all $\bSigma_{\beps_{ij}}$ by $\bI_{3\times3}$, or \cref{eq:LOST_Ax=b} replacing all $q_j$ by a constant. Therefore, we observe that the optimal solution will differ from that of the DLT when the measurement range to the 3D point, or the 2D noise, varies between different views.

In the rest of the paper, we refer to \texttt{LOST} as the algorithm that solves \cref{eq:LOST_Ax=b}, \ie that only accounts for 2D uncertainties. We refer to \texttt{LOSTU} as the algorithm that solves \cref{eq:LOST_Normal}, \ie accounting for uncertainties of camera parameters and 2D noise. Neither \texttt{LOST} or \texttt{LOSTU} use an \emph{a-priori} estimate of $\bX_i$, as they find the range with \cref{eq:approx_range}. The algorithm \texttt{DLT} refers to \cref{eq:LOST_Ax=b} where all $q_j = 1$. The covariance of a point --- accounting for both 2D and camera uncertainties --- triangulated with linear triangulation methods like \texttt{DLT}, \texttt{LOST}, and \texttt{LOSTU} can be found at negligible computational expense, and expressions are in Ref.~\cite{Henry:2022}.

\subsection{Optimal camera center estimation} \label{sec:LOST_resection}
The underlying analysis for resection (\eg in navigation) is identical to the one in \cref{sec:LOST_intersection}. Consider the cost function
\begin{equation} \label{eq:cost_function_camcenter}
    J(\bc_j) = \sum_{i\in\mathcal{V}_j} \beps_{ij}^T \bSigma_{\beps_{ij}}^{\dagger} \beps_{ij},
\end{equation}
Then the optimal camera center $\bc_j$ is found with
\begin{equation}
    \begin{aligned}
    \left(\sum_{i \in \mathcal{V}_j} \bR^{C_j}_W \left[ \bK_j^{-1} \bx_{ij} \times \right] \bSigma_{\beps_{ij}}^{\dagger} \left[ \bK_j^{-1} \bx_{ij} \times \right] \bR^W_{C_j} \right) \bc_j \\
    = \sum_{i \in \mathcal{V}_j} \bR^{C_j}_W \left[ \bK_j^{-1} \bx_{ij} \times \right] \bSigma_{\beps_{ij}}^{\dagger} \left[ \bK_j^{-1} \bx_{ij} \times \right] \bR^W_{C_j} \bX_i .
\end{aligned}
\end{equation}

\subsection{Optimality of midpoint} \label{sec:midpointoptimal}
Some experiments have found that midpoint performs well when triangulating with camera pose noise \cite{Nasiri:2023}, but little explanation is provided to rationalize those results. Starting from \cref{eq:LOST_Normal}, we show that the midpoint is the optimal method when, 1) all cameras have the same position covariance, and 2) this camera position uncertainty dominates all other error sources. In this case, using \cref{eq:partial_translation} and \cref{eq:residual_cov}, the covariance of $\beps_{ij}$ is 
\begin{equation}
    \bSigma_{\beps_{ij}} = - \left[ \bK_j^{-1} \bx_{ij} \times \right] \bR^W_{C_j} \bSigma_{\bc_j} \bR^{C_j}_{W} \left[ \bK_j^{-1} \bx_{ij} \times \right],
\end{equation}
where we considered the fact that $[\, \cdot \, \times]^T = -[\, \cdot \, \times]$.
Assuming an isotropic camera center noise under the form $\bSigma_{\bc_j} = \bI$, and noting $\bR^W_{C_j} \bR^{C_j}_{W} = \bI$ , we compute its pseudo-inverse as
\begin{equation}
    \bSigma_{\beps_{ij}}^{\dagger} = -\frac{1}{\| \bK_j^{-1} \bx_{ij} \|^4} \left[ \bK_j^{-1} \bx_{ij} \times \right]^2.
\end{equation}
It follows that
\begin{equation}
    \bR^{C_j}_W \left[ \bK_j^{-1} \bx_{ij} \times \right] \bSigma_{\beps_{ij}}^{\dagger} \left[ \bK_j^{-1} \bx_{ij} \times \right] \bR^W_{C_j} = \bR^{C_j}_W \left[ \ba_{ij} \times \right]^4 \bR^W_{C_j}.
\end{equation}
Note that $\left[ \ba_{ij} \times \right]^4 = \bI - \ba_{ij} \ba_{ij}^T$, so we rewrite \cref{eq:LOST_Normal} as 
\begin{equation}
        \begin{aligned} \label{eq:midpoint}
    \left(\sum_{j \in \mathcal{T}_i} \bR^{C_j}_W (\bI - \ba_{ij} \ba_{ij}^T) \bR^W_{C_j} \right) \bX_i \\
    = \sum_{j \in \mathcal{T}_i} \bR^{C_j}_W (\bI - \ba_{ij} \ba_{ij}^T) \bR^W_{C_j} \bc_j 
    \end{aligned}
\end{equation}
which is precisely the formulation for the midpoint triangulation in $n$-view \cite{szeliski:2022_coputervision}. The proof for the resection case is analogous. We still note that \texttt{LOSTU} is a more general framework that can treat camera position uncertainties of different amplitudes alongside angular noise. Thereafter, \texttt{midpoint} refers to the algorithm solving \cref{eq:midpoint}. Comparing \cref{eq:midpoint} with \cref{eq:LOST_Normal}, we observe that \texttt{midpoint} is nothing else than the \texttt{DLT} with unit normalized LOS.

\section{Triangulation in SfM framework}
Structure from Motion (SfM) is the process of reconstructing a 3D scene from  2D images. Today's SfM pipelines have matured considerably since early work \cite{Sutherland:1974_3Ddata, longuet-higgins:1981_reconstructing}, and are now routinely used to reconstruct large urban scenes \cite{Snavely:2006_PhotoTourism, snavely:2008_modeling, Argawal:2011_RomeInADay, Schonberger:2016_SfMRevisited}, terrain \cite{Westoby:2012_SfMGeoscience}, and celestial bodies \cite{Palmer:2022_SPC, Driver:2023_AstroVision}. This section is aimed at pointing out the particular aspects to consider when triangulating in SfM.

There are many different approaches when it comes to SfM, but all have in common that features need to be extracted from---and matched between--- images. This process can be done by well-known handcrafted algorithms \cite{Lowe:1999_SIFT, lowe:2004_SIFT, BAY:2008_SURF, Rublee:2011_ORB} or learning algorithms \cite{DeTone:2018_SuperPoint, Revaud:2019_R2D2, Luo:2020_ASLFeat}. In some cases the 2D uncertainty that comes with those features can be rigorously estimated, though it is also common to simply assume a fixed value (\eg, 1 pixel). From here, we often categorize distinct approaches.
\paragraph{Sequential SfM}
The extracted features are used to estimate an initial relative pose between at least two starting views. This seeding process is preferably done in a dense central place \cite{Haner:2012_nextBestView}. This process can be done with the five point algorithm \cite{Nister:2004_fivepoint} for calibrated cameras or the eight point algorithm \cite{Hartley:1997_eightpoint} for uncalibrated cameras, which are often coupled with an outlier detection scheme like RANSAC \cite{Fischler:1981_RANSAC}. The 3D points commonly seen by the cameras can be triangulated. Then, an initial bundle adjustment (BA) allows one to obtain the initial covariance of the poses and structure \cite{hartley_zisserman_2004, Triggs:2000_BA}.

Views are then added sequentially. Several options exist when it comes to selecting the next best view. It can involve propagating the covariance and selecting the best camera, but this is slow in practice since it requires many unnecessary view estimations. Instead, simple rules like choosing the camera that sees the most points work well in practice \cite{Haner:2012_nextBestView}. The estimation of a view pose using 3D points is referred to as the \emph{Perspective-n-Point} (PnP). The PnP problem can be solved using from 3 \cite{Grunert:1841, Dhome:1989, Gao:2003} to $n$ \cite{lepetit:2009_epnp, hartley_zisserman_2004, Zheng:2013_opnp, kneip:2014_upnp} measurements, and the reprojection error is often the quantity minimized, but these break the pattern of maximum-likelihood since not all uncertainties are taken into account. Other works propose to leverage uncertainties from the 3D reconstructed points directly in the PnP \eg \cite{ferraz:2014, Urban:2016, Vakhitov:2021_EPnPU, jahani:2023_AQPnP}. The view pose uncertainties arising from a process like the one in Ref.~\cite{Vakhitov:2021_EPnPU} can be obtained because they use a subsequent iterative refinement with Levenberg-Marquardt (LM) or Gauss-Newton (GN).

Once a camera is added, all points seen by two views or more are triangulated. The choice of the best triangulation method depends on the magnitude of the pose noise and availability/accuracy of the pose covariance, and the triangulation covariance expressions \cite{Henry:2022} can help in the analysis. 

Then, either a BA step is performed, or the next view is estimated. Careful implementations can greatly reduce the computational load of BA \cite{Schonberger:2016_SfMRevisited, Wu:2013_LinearTimeSfM}. There are often additional triangulation steps before and after BA \cite{Schonberger:2016_SfMRevisited}. Depending on the reconstruction, obtaining the covariance of the poses after BA may be practically difficult. If the covariance are not obtainable, BA still reduces the errors of the poses and thus may render $L_2$ triangulation more profitable.

\paragraph{Global SfM}
Sequential SfM, although very accurate, can be slow. Popular and faster alternatives are global SfM \cite{Moulon:2013_GlobalSfM, sweeney:2015_theia, Chen:2021_RotationAveraging} that estimate all poses first, triangulate, and then perform BA only once. After estimating relative poses with aforementioned tools, the \emph{averaging} process refines them into a coherent set \cite{Hartley:2013_RA, Chatterjee:2013_RotationAveraging, Dellaert:2020_ShonanRA, Chen:2021_RotationAveraging}. Ref.~\cite{Zhang:2023_RevisitingRotationAveraging} propagates uncertainties into weights for estimated rotations. This may be particularly interesting to use in conjunction with \texttt{LOSTU}.

\paragraph{Hybrid SfM}
A proposed middle-ground to sequential or global SfM is \emph{hybrid} SfM \cite{Cui:2017_HybridSfM, Dennison:2023_RISfM}, where all camera rotations are first estimated, and the sequential part only focuses on camera center and structure reconstruction. In this case, the problem reduces to a sequence of triangulation problems. The LOST framework allows to seamlessly obtain the optimal camera center using the 3D structure as in \cref{sec:LOST_intersection}, and 3D structure using camera centers as in \cref{sec:LOST_resection}.

\paragraph{SLAM}
In Simultaneous Localization and Mapping (SLAM), the emphasis is placed on both the structure and the position of the observer \cite{Campos:2021_ORBSLAM3}. It is usually done in an incremental manner with more emphasis in real-time application. Covariances may be readily available when working with Kalman Filters \cite{Davison:2007_MonoSLAM}. Furthermore, optimization frameworks like iSAM \cite{Kaess:2008_iSAM, Kaess:2012_iSAM2} may be utilized to quickly estimate the marginal covariances.

\section{Triangulation experiments}
\subsection{Two-view triangulation}
\begin{figure*}[t]
  \centering
  \begin{subfigure}{0.23\linewidth}
    \includegraphics[width=1\linewidth]{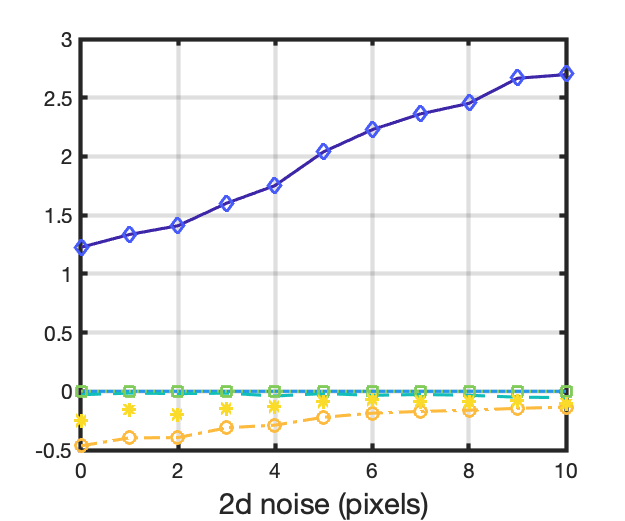}
    \caption{Variation in pixel noise.}
    \label{fig:2v_var-pixelnoise}
  \end{subfigure}
  \hfill
  \begin{subfigure}{0.23\linewidth}
    \includegraphics[width=1\linewidth]{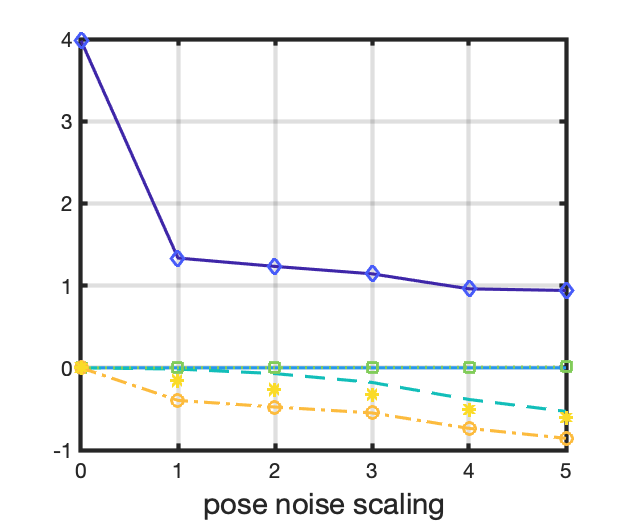}
    \caption{Variation in pose uncertainty.}
    \label{fig:2v_var-posenoise}
  \end{subfigure}
  \hfill
  \begin{subfigure}{0.23\linewidth}
    \includegraphics[width=1\linewidth]{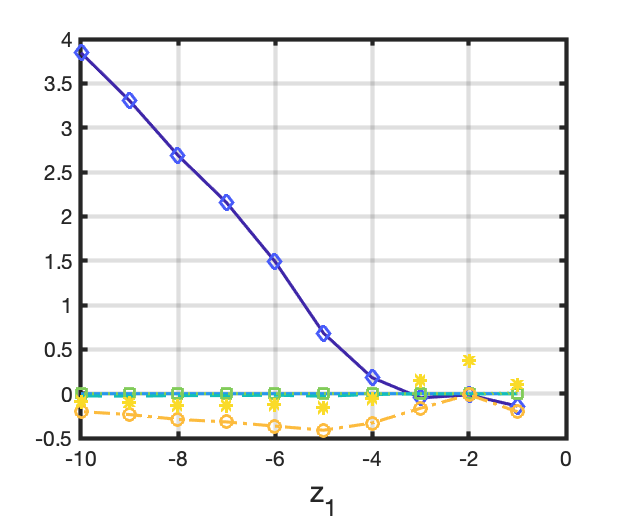}
    \caption{Variation of $z_1$ (depth).}
    \label{fig:2v_var-depth}
  \end{subfigure}
  \hfill
    \begin{subfigure}{0.23\linewidth}
    \includegraphics[width=1\linewidth]{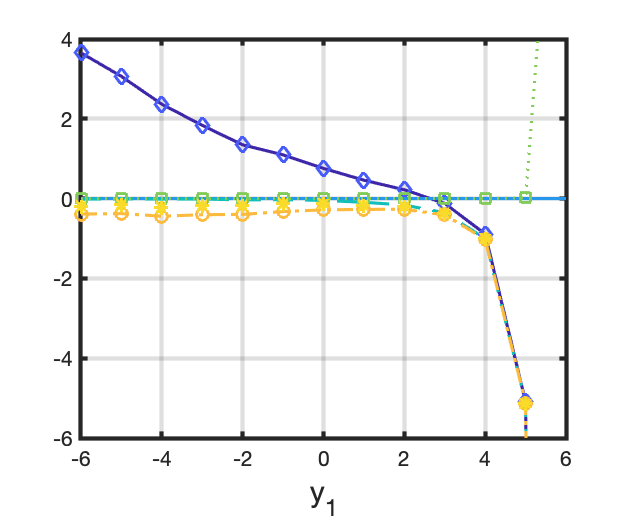}
    \caption{Variation of $y_1$ (side).}
    \label{fig:2v_var-side}
  \end{subfigure}
  \begin{subfigure}{1.\linewidth}
  \centering
    \includegraphics[width=0.7\linewidth]{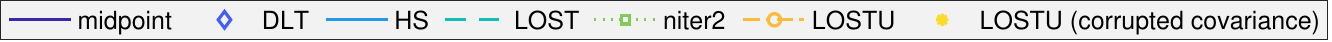}
    % \caption{variation of pose noise.}
    \label{fig:2v_legends}
  \end{subfigure}
  \caption{Percentage of RMSE deterioration with respect to \texttt{HS} for two view triangulation. Negative value means better than minimizing the reprojection error. Curves for \texttt{DLT} and \texttt{midpoint} overlap. We observe a clear gap between optimal and suboptimal methods.}
  \label{fig:2v}
\end{figure*}

Many applications in 3D vision and navigation use a limited number of cameras. The case of solving 2-view triangulation has been widely discussed in the literature \cite{Hartley:1997, Kanatani:2008, Lindstrom:2010}. In this section, we compare the performance of \texttt{LOST} against the 6th order polynomial of Hartley and Sturm (\texttt{HS} \cite{Hartley:1997}), the fast optimal solution developed by Lindstrom (\texttt{niter2} \cite{Lindstrom:2010}), \texttt{DLT}, \texttt{midpoint}, and \texttt{LOSTU}, the only solution here that takes into account camera pose uncertainties. For Robustness analysis, we also show a version where the pose uncertainties are fed to \texttt{LOSTU} with wrong values. The corrupted covariances are made with a factor between 1/2 and 2, independently random for each camera, pose covariance, and rotation covariance. The purpose of this experiment is double. First, it shows that directly solving the simple \texttt{LOST} linear system gives essentially the same result as the more complicated (and slower) Hartley and Sturm polynomial. Second, it aims to address the findings in \cite{lee:2019triangulation, Nasiri:2023} that tested optimal triangulation in geometries where optimal triangulation simply has no advantage.

Suppose a 3D object is placed at the origin and observed by two cameras that are placed in $ c_1 = [0, y_1, z_1]^T $ and $ c_2 = [0, 2, -2]^T$. In the nominal case, camera one is placed such that $y_1 = -2$, $z_1 = -6$. The camera has an effective focal length of $400$. We assume a 2D noise of $\sigma_{px} = 1$ pixel, a camera rotation noise of $\sigma_{\phi} = 0.5$ deg, and a camera center noise of $\sigma_{\bc} = 0.03$.  The cameras are rotated such that they point towards the 3D point (orbital configuration). Independently, we vary $\sigma_x$, the camera pose, $y_1$, and $z_1$. To make the results more graphically meaningful, we show the results in \cref{fig:2v} in percentage of position RMSE compared to the HS polynomial solution:
\begin{align}
    \text{Error} = 100 \times \left(\text{RMSE}_\tau - \text{RMSE}_\text{HS}\right)/\text{RMSE}_\text{HS},
\end{align}
where $\tau$ is the tested triangulation method. \cref{fig:2v_var-pixelnoise} highlights the gap in performance between the optimal solutions and \texttt{midpoint} or \texttt{DLT}. 

The fact that a performance gap persists when there is no pixel noise is due to the remaining uncertainty in the camera pose. In \cref{fig:2v_var-posenoise}, we observe that the non-optimal algorithms get better relative to the polynomial solution of Hartley and Sturm as the camera pose uncertainties increase. We also observe that \texttt{LOST} behaves comparatively better in that case, and \texttt{LOSTU} always has the lowest error. When there are no camera pose uncertainties, all optimal methods have the same RMSE. \cref{fig:2v_var-depth} confirms the analysis of \cref{eq:LOST_weight} in that the the sub-optimal methods behave similarly to the optimal methods at $z_1=-2$, when the geometry is symmetric and the ranges are all the same. As the depth decreases, the relative importance of the camera center noise increases and \texttt{midpoint} performs well. Finally, \cref{fig:2v_var-side} shows what happens when the angle between the two observations progressively decrease. In this case we observe that the classical two view optimal solutions, \texttt{HS} and \texttt{niter2}, start to comparatively lose in quality. This trend is less pronounced if no camera noise is added. In our experiments, \texttt{LOST} and \texttt{LOSTU} exhibit better behavior in low parallax.

%This prompts us to investigate the low parallax behavior of \texttt{LOST} compared to standard methods like \texttt{HS}. This experiment is done without camera pose noise. Set $c_1 = [1, 0, -6]$ and $c_2 = [1, 0, -5]$, such that the inter line-of-sight angle is less than 2 degree. As pixel noise increases, both \texttt{HS} and \texttt{niter2} become unstable, whereas \texttt{LOST} stays on par with the \texttt{DLT} and the \texttt{midpoint} methods. The results for this experiment and a standard deviation of 1 pixel is available in \cref{tab:low_plx}. Other two-view triangulation techniques can be considered when low parallax is the main focus \cite{lee:2019triangulation}.

The runtimes of the tested algorithms are presented in \cref{tab:2v_runtimes}. The experiment has been performed with MATLAB and a 2.3 GHz Intel Core i9. Our MATLAB implementation of the \texttt{niter2} by Lindstrom \cite{Lindstrom:2010} remains the fastest method to triangulate optimally in two view. \texttt{LOST} is still twice as fast as solving the \texttt{HS} and comes with additional robustness and scalability benefits.

This experiment shows that optimal triangulation can still lead to improvements over non-optimal schemes, even when substantial camera pose uncertainties exist. It does show operational regions where standard optimal triangulation algorithms become less stable. The LOST algorithm gives a similar solution to standard $L_2$ triangulation in nominal cases, while also responding better to camera noise and low parallax.

\begin{table}[h!]
    \centering
    \caption{Mean runtimes of different triangulation methods, in $\mu$s.}
    \begin{tabular}{cccccc}
    \toprule
        \texttt{midpoint} & \texttt{DLT} & \texttt{LOST} & \texttt{HS} & \texttt{niter2} & \texttt{LOSTU} \\
        \midrule
        12.8 & 19.9 & 23.5 & 52.4 & 12.9 & 43.7\\
    \bottomrule
    \end{tabular}
    \label{tab:2v_runtimes}
\end{table}

\subsection{N-view triangulation}
This experiment aims to compare multiple triangulation solutions in a typical SfM geometry. The algorithms selected are \texttt{midpoint}, \texttt{DLT}, \texttt{LOST} as the $L_2$ optimal triangulation, and \texttt{LOSTU} as the triangulation accounting for all uncertainties. The DLT method is also implemented with a refinement using factor graphs and LM \cite{Dellaert2012FactorGA} to minimize either the reprojection error with \texttt{DLT+LM (reproj)}, or the Mahalanobis distance \texttt{DLT+LM (Mahalanobis)}. For comparison, a DLT solution that is refined by LOSTU is also added, \texttt{DLT+LOSTU}.

A single point placed at $\bX = [2, 1, 0 ]$ is triangulated by $m=50$ cameras randomly spawned in a domain $\mathcal{D}_{\text{cam}} = \{[x_{\min}, x_{\max}] = [-10, 10], [y_{\min}, y_{\max}] = [-10, 10], [z_{\min}, z_{\max}] = [-50, -10]\}$. Each camera is oriented to look in the $+z$ direction with a random deviation of $2$deg. The effective focal length of the camera is 800 and an isotropic 2D noise of 1 pixel is added to the measurements. Furthermore, a camera rotation noise of $\sigma_{\phi} = 0.05$deg and translation of $\sigma_{\bc} = 0.02$ is added for every camera. These pose uncertainties are randomly scaled by a factor from 1/2 to 2 for each camera, so that all cameras have different uncertainties.

\begin{figure*}
  \centering
  \begin{subfigure}{0.23\linewidth}
    \includegraphics[width=1\linewidth]{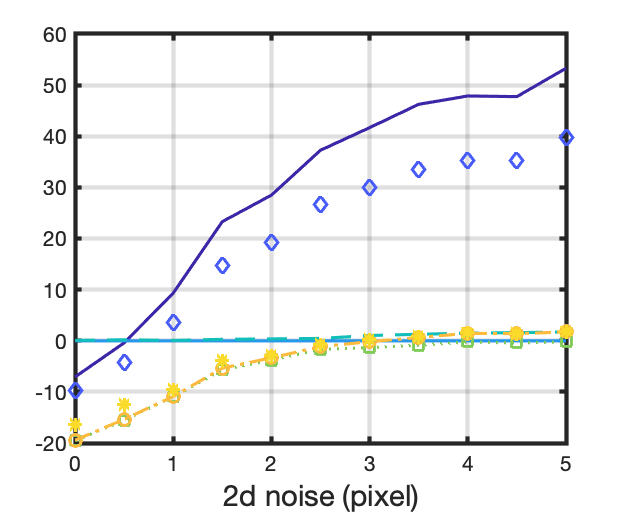}
    \caption{}
    \label{fig:nv_var-pixelnoise}
  \end{subfigure}
  \hfill
  \begin{subfigure}{0.23\linewidth}
    \includegraphics[width=1\linewidth]{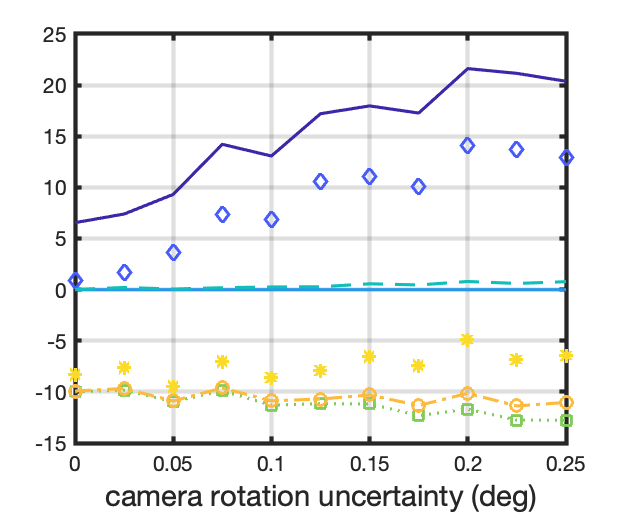}
    \caption{}
    \label{fig:nv_var-rot}
  \end{subfigure}
  \hfill
    \begin{subfigure}{0.23\linewidth}
    \includegraphics[width=1\linewidth]{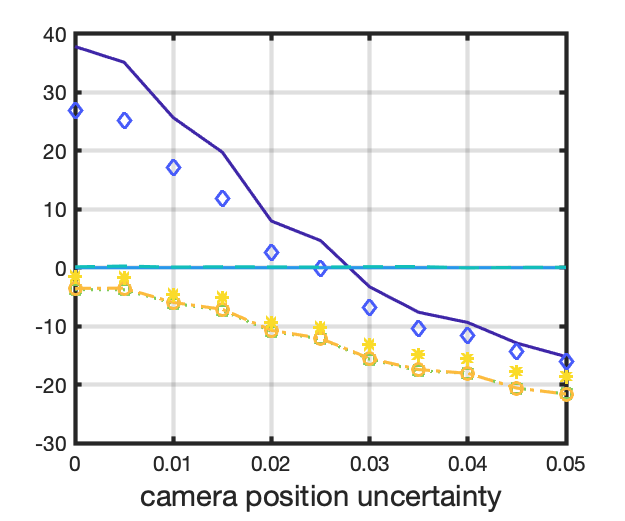}
    \caption{}
    \label{fig:nv_var-pos}
  \end{subfigure}
  \hfill
    \begin{subfigure}{0.23\linewidth}
    \includegraphics[width=1\linewidth]{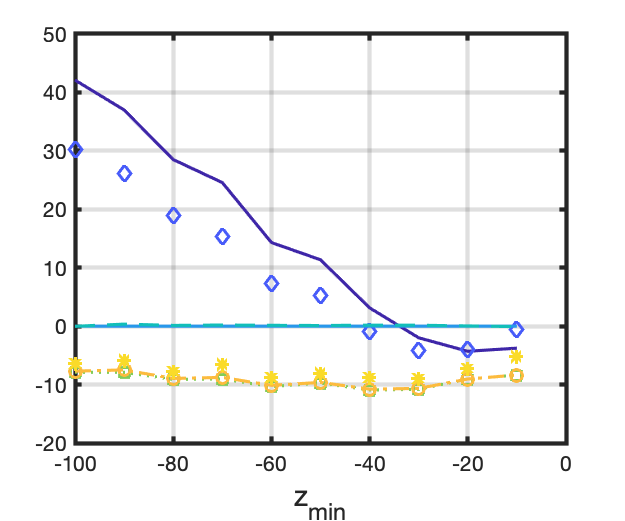}
    \caption{}
    \label{fig:nv_var-depth}
    \end{subfigure}
  \begin{subfigure}{1.\linewidth}
  \centering
    \includegraphics[width=0.8\linewidth]{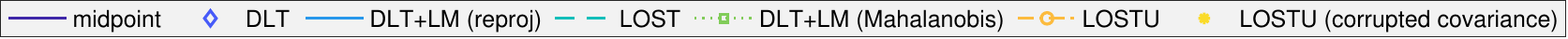}
    \label{fig:nv_legends}
  \end{subfigure}
  \caption{Percentage of RMSE deterioration with respect to \texttt{DLT+LM (reproj)} for 50-view triangulation. Negative value means better than minimizing the reprojection error. }
  \label{fig:n-view}
\end{figure*}
\begin{figure}
\begin{center}
    \begin{subfigure}{0.23\linewidth}
    \includegraphics[width=1\linewidth]{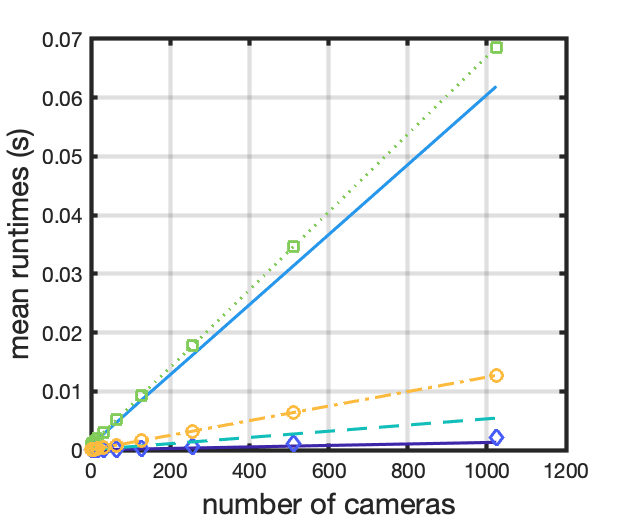}
    \caption{}
    \label{fig:ncamruntimes}
    \end{subfigure}
    \begin{subfigure}{0.23\linewidth}
    \includegraphics[width=1\linewidth]{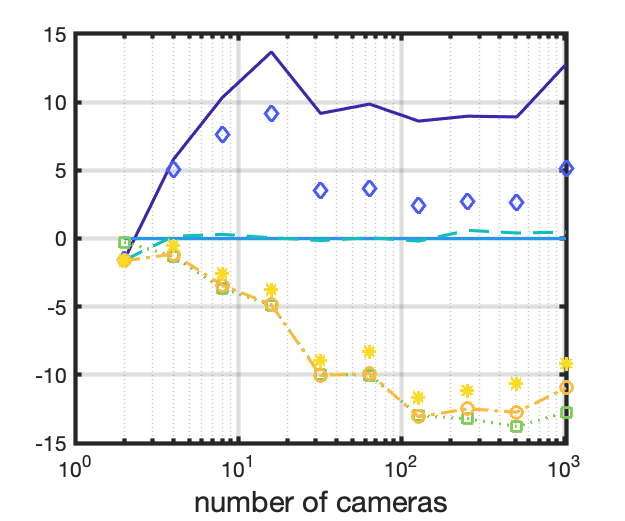}
    \caption{}
    \label{fig:ncamRMSE}
    \end{subfigure}
\end{center}
    \caption{Evolution of triangulation as the number of views increases. (a): All techniques exhibit linear runtime complexity but have different slopes. (b) Percentage of RMSE deterioration with respect to \texttt{DLT+LM (reproj)}. Legends can be found in \cref{fig:n-view}}
    \label{fig:n-view-var-ncam}
\end{figure}

Each of the parameters will be varied independently to study their effects on classical triangulation solutions. For each set of parameters, we perform triangulation 5000 times, where camera poses and measurements are regenerated at each iteration. The position RMSE are recorded in \cref{fig:n-view} while the mean runtimes are displayed in \cref{fig:n-view-var-ncam}.

One can observe that \texttt{midpoint} is the fastest method but it provides a suboptimal solution, except when the position uncertainty of the cameras dominate, which confirms the results in \cref{sec:midpointoptimal}. The \texttt{midpoint} will not coincide with the optimal solution for high camera center noise because cameras have different pose noise. The \texttt{DLT} is slower, but still time-efficient, and it is consistently better than the midpoint in this experiment. Minimizing the reprojection error is still a strategy that yields significant improvements over \texttt{DLT} and \texttt{midpoint}, depending on multiple factors. First, the geometries favouring greater variations in distance between the views give an edge to $L_2$ triangulation. Second, noise in the position of the camera centers reduce the relative performance of $L_2$ triangulation. Third, the number of views may change the relative performance between methods. In our findings, a moderate number of views tend to favor minimizing the reprojection error, and then the relative performance may go in either direction depending on the 2D noise to pose noise ratio. \texttt{LOST} performs statistically identically to the LM refinement at a fraction of the computational cost. This behavior slightly deviates for a high number of views when the angular noise becomes significant, probably due to the fact that the weighting is done by estimating the range using noisy measurements.

All the methods above do not require any knowledge of the covariance matrices. When these are available, the methods that properly account for them are always statistically better, sometimes substantially so. \texttt{LOSTU} again performs equivalently to the LM refinement, and can be slightly more robust if used in a refinement way. As a final implementation tip, \texttt{LOSTU} can be made as fast as \texttt{LOST} if the residual covariance matrix in \cref{eq:residual_cov} is approximated by a diagonal matrix to speed-up the computation of the pseudo-inverse. In that case the results were found to still be close to the standard \texttt{LOSTU}.

\section{Sequential reconstruction example}
Reference \cite{Haner:2012_nextBestView} proposed a reconstruction pipeline where cameras and points are sequentially added and covariance propagated throughout. In their approach, cameras are iteratively refined to the maximum likelihood estimate when they are estimated for the first time. Cameras are not accepted if their reprojection error is too large. Points seen by at least two views are triangulated and iteratively refined to a maximum likelihood estimate. Similarly, points are not accepted if their reprojection error and covariance are too large. While BA often remains necessary for the most accurate reconstructions, this way of reconstructing has been shown to yield decent sequential reconstructions without it. We choose to follow a similar setup to build a reconstruction problem where the camera poses have an estimated covariance. This allows us to then compare the performance of different triangulation solutions.

For camera pose estimation, we distinguish cases where the reprojection error is minimized with EPnP \cite{lepetit:2009_epnp} (and the covariance does \emph{not} correctly account for all terms) vs. the case where 3D point uncertainties are also taken into account in EPnPU \cite{Vakhitov:2021_EPnPU} (the covariance is correctly propagated). We test the following configurations:  1) \texttt{EPnP+DLT} (standard practice), 2) \texttt{EPnP+LOST} (minimized reprojection error), 3) \texttt{EPnPU+DLT} (correct covariance propagation), 4) \texttt{EPnPU+midpoint} (correct covariance propagation), and 5) \texttt{EPnPU+LOSTU} (correct covariance propagation and maximum likelihood solution).

\subsection{ETH3D}

\begin{table*}[t!]
    \centering
    % NiceTabular
    \caption{Sequential reconstruction of ETH3D high-res datasets \cite{schoeps:2017_ETH3D}, where reconstruction scores have been computed using the standardized evaluation tool in~\cite{schoeps:2017_ETH3D}, with a tolerance of $10cm$, where $a$=accuracy and $c$=completeness. The SfM reconstruction from \cite{schoeps:2017_ETH3D} is put as a reference traditional SfM.}
    \resizebox{\linewidth}{!}{
    \begin{NiceTabular}{llcccccc>{\columncolor[gray]{0.9}}c}[colortbl-like]
    \toprule
        Scene & Algorithm & estimated & estimated & time (s) & reprojection & \multicolumn{3}{c}{3D reconstruction scores} \\ \cline{7-9}
        & & points & views & & error (pixels) & $a$ (\%) & $c$ (\%) & $F_1$ (\%) \\
        \midrule
         \cellcolor[gray]{0.9}delivery\_area & \cellcolor[gray]{0.9}SfM from \cite{schoeps:2017_ETH3D} & \cellcolor[gray]{0.9}31,978 & \cellcolor[gray]{0.9}44 & \cellcolor[gray]{0.9}- & \cellcolor[gray]{0.9}0.8831 & \cellcolor[gray]{0.9}92.88 &  \cellcolor[gray]{0.9}23.70 & \cellcolor[gray]{0.9}37.77  \\
         &\texttt{EPnP+DLT} & 30,704 & 43 & 20 & 0.9277 & 88.15 & 22.17 & 35.44 \\
         &\texttt{EPnP+LOST} & 30,858 & 43 & 19 & 0.9270 & 87.55 & 22.09 & 35.29 \\
         &\texttt{EPnPU+midpoint} & 30,641 & 44 & 24 & 0.9392 & 91.15 & 22.97 & 36.69 \\
         &\texttt{EPnPU+DLT} &  30,864 & 44 & 21 &  0.9285 & 91.45 & 23.08 & 36.87 \\
         &\texttt{EPnPU+LOSTU} & 30,964 & 44 & 23 &  0.9201 & \textbf{92.45} & \textbf{23.37} & \textbf{37.30} \\
         \midrule
         \cellcolor[gray]{0.9} terrains & \cellcolor[gray]{0.9} SfM from \cite{schoeps:2017_ETH3D} & \cellcolor[gray]{0.9} 18,553 & \cellcolor[gray]{0.9} 42 & \cellcolor[gray]{0.9} - & \cellcolor[gray]{0.9} 0.8886 & \cellcolor[gray]{0.9} 95.84 & \cellcolor[gray]{0.9} 24.15 & \cellcolor[gray]{0.9} 38.59 \\
         &\texttt{EPnPU+midpoint} &  14,799 & 30 & 12 & 1.0861 & 96.03 & 19.90 & 32.96 \\
         &\texttt{EPnPU+DLT} & 14,813 & 30 & 13 & 1.1100 & 96.01 & 19.81 & 32.85 \\
         &\texttt{EPnPU+LOSTU} & 16,671 & 36 & 12 & 0.9181 & \textbf{96.55} & \textbf{20.65} & \textbf{34.01} \\
         \midrule
         \cellcolor[gray]{0.9} facade & \cellcolor[gray]{0.9} SfM from \cite{schoeps:2017_ETH3D} & \cellcolor[gray]{0.9} 85,096 & \cellcolor[gray]{0.9} 76 & \cellcolor[gray]{0.9} - & \cellcolor[gray]{0.9} 0.9914 & \cellcolor[gray]{0.9} 84.43 & \cellcolor[gray]{0.9} 27.15 & \cellcolor[gray]{0.9} 41.09  \\
         &\texttt{EPnPU+midpoint} & 79,634 & 75 & 114 & 0.9700 & 84.93 & 15.46 &  26.15 \\
         &\texttt{EPnPU+DLT} & 80,948 & 75 & 123 & 0.9663 & 83.78 & 14.71 &  25.03 \\
         &\texttt{EPnPU+LOSTU} & 82,089 & 75 & 129 & 0.9590 & \textbf{85.92} & \textbf{18.18} &  \textbf{30.00} \\
         \bottomrule
    \end{NiceTabular}}
    \label{tab:SfM_results}
\end{table*}

ETH3D \cite{schoeps:2017_ETH3D} is a dataset that offers various multiview scenes with raw images, or already matched features. We select it particularly because it contains very high-precision ground truth obtained with scanners. We start from the features already matched in the dataset. Starting from two views with around 700 common points, we estimate their relative pose along with a RANSAC scheme to find inliers. Inliers are then triangulated and the initial pose uncertainty is estimated with BA. The next best view is chosen as the camera that observes the most estimated points. If the reconstructed camera has a mean reprojection error higher than 5 pixels, then it is not considered until 3D points are better refined, and another camera is estimated instead. Points are re-triangulated as more views are added. 3D points whose reprojection error is higher than 5 pixels, or standard deviation $\sigma_X = \sqrt{\text{trace}(\bSigma_X)}$ exceeds about 0.2 m (scale assumed known) are not accepted. An initial 2D standard deviation of 1 pixel is assumed to compute the initial covariances---this value is only a guess of the true value. While the reprojection error is not the metric to minimize for maximum likelihood, it is still a convenient representation of the global condition of the reconstruction that does not require covariance to be computed.

We show the results on three of the high-res scenes in \cref{tab:SfM_results}, where we compare our sequential reconstruction without BA (outside of the initial geometry) to the SfM from the dataset Ref.~\cite{schoeps:2017_ETH3D}. We observe that the solutions that do not properly propagate the covariance exhibit worse reconstruction metrics. Since no BA is performed after the initial geometry in this experiment, camera pose noise remains high and minimizing the reprojection error in the triangulation step does not coincide with a maximum likelihood estimate anymore. This explains why \texttt{EPnP+LOST} does not perform better than \texttt{EPnP+DLT}. When covariance propagation is properly taken into account, \texttt{LOSTU} consistently triangulated more points than the \texttt{DLT} and \texttt{midpoint}, while simultaneously fitting closer to the scanner ground truth. The accuracy of \texttt{EPnPU+LOSTU} was often better than the accuracy of the reference SfM.

Overall, this experiment shows that correct estimation and propagation of the covariance in triangulation can lead to results that are closer the reconstructed structure. Depending on the required fidelity of the reconstruction, this can lead to a simple SfM pipeline with reduced need for BA. These conclusions are similar to those found in Ref.~\cite{Haner:2012_nextBestView}.
\begin{figure}[t!]

\hfill \begin{subfigure}{0.45\linewidth}
    \includegraphics[width=0.8\linewidth]{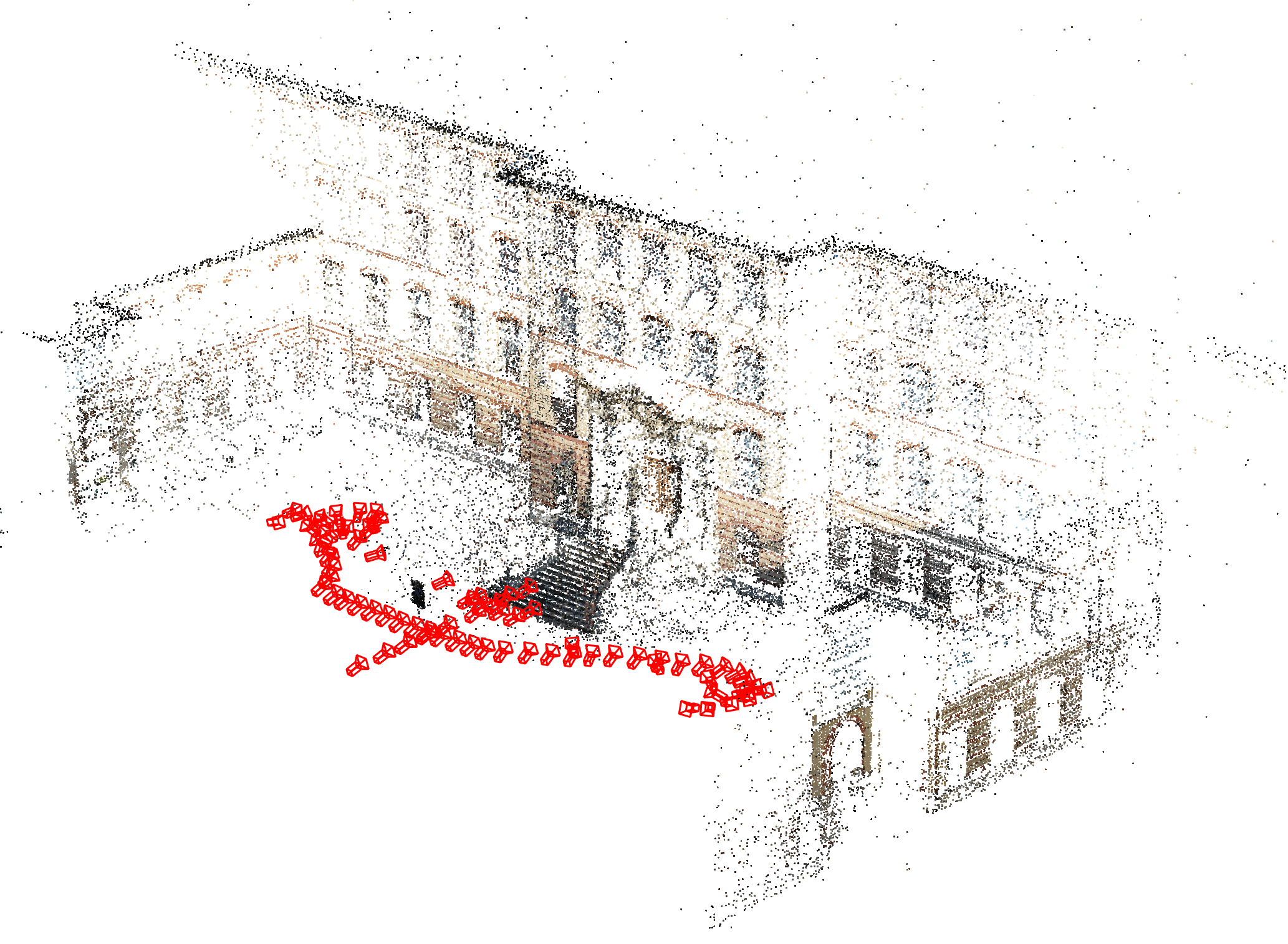}
\end{subfigure}
\begin{subfigure}{0.45\linewidth}
    \includegraphics[width=0.8\linewidth]{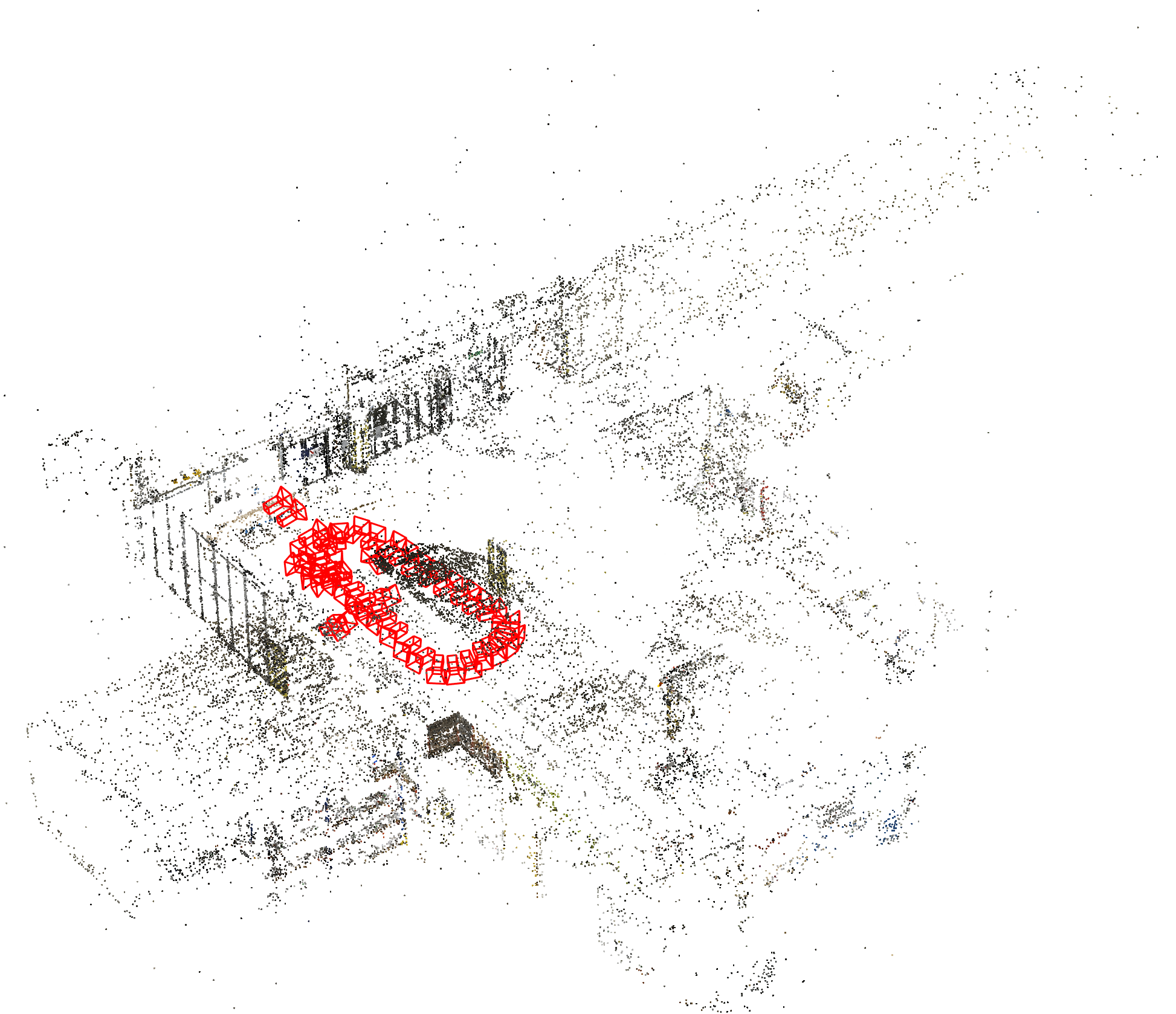}
\end{subfigure}
  \caption{\texttt{EPnPU+LOSTU} reconstruction for facade (left) and delivery\_area (right) \cite{schoeps:2017_ETH3D}.}
  \label{fig:facade}
\end{figure}

\subsection{Vesta reconstruction}
The Astrovision dataset \cite{Driver:2023_AstroVision} offers the possibility to reconstruct several asteroids. We choose the ASLFeat features \cite{Luo:2020_ASLFeat} trained on Astrovision data, \emph{ASLFeat-CVGBEDTRPJMU}, since these were shown to extract a large number of features with good precision on asteroid images. After matching and outlier rejection, we perform the sequential SfM to 3D reconstruct Vesta. The maximum reprojection error to accept a camera or a point is set to 5 pixels. The maximum covariance to accept a 3D point is set to a value such that the reconstructed surface looks arbitrarily smooth. Results for \texttt{EPnPU+DLT} versus \texttt{EPnPU+LOSTU} are found in \cref{tab:SfM_results}, and the visual in \cref{fig:vesta_reconstruction}, in which we observe that \texttt{LOSTU} estimated more than double the number of cameras and points compared to \texttt{DLT}. The mean reprojection and reconstruction errors are lower in the case of \texttt{LOSTU}.

\begin{table}[h!]
    \centering
    \caption{Reconstruction metrics for the RC3b segment of Vesta.}
    \begin{tabular}{lccc}
    \toprule
    Algorithm & points & views & rep. err. \\
    \midrule
    ground truth+\texttt{LOST} & 37,205 & 65 & 0.768 \\
    \texttt{EPnPU+DLT} & 10,396 & 24 & 1.104 \\
    \texttt{EPnPU+LOSTU} & 28,865 & 58 & 0.929 \\
    \bottomrule
    \end{tabular}
    \label{tab:Vesta}
\end{table}

\section{Conclusion}
This paper presents a light and complete framework to triangulate with camera-parameter uncertainties. Geometry, the number of views, and the relative weight of camera parameters to 2D noise all play a crucial role when choosing the right triangulation method. When information on uncertainties is available, \texttt{LOSTU} can lead to substantial improvements in reconstruction scenarios. However, estimating accurate pose covariances from images can be challenging, and there remains room for improvement in how easily these could be obtained.

\section{Acknowledgements}
We thank Travis Driver for his help in the asteroid reconstruction example and valuable feedback on this manuscript. We also thank Michael Krause and Priyal Soni for their thoughtful comments.

\bibliographystyle{splncs04}
\bibliography{main}
\end{document}